\documentclass[runningheads]{llncs}
\usepackage{graphicx}
\usepackage{bbding}
\usepackage{amsmath}
\usepackage{multirow}
\usepackage{tabulary}
\begin{document}
\title{EndoFinder: Online Image Retrieval for Explainable Colorectal Polyp Diagnosis}
\titlerunning{Image Retrieval for Polyp Diagnosis}
\author{Ruijie Yang\inst{1,2,3}\dag \and Yan Zhu\inst{4,5}\dag \and Peiyao Fu\inst{4,5}, Yizhe Zhang\inst{6} \and Zhihua Wang\inst{3}\Envelope \and Quanlin Li\inst{4,5} \and Pinghong Zhou\inst{4,5} \and Xian Yang\inst{7} \and Shuo Wang\inst{1,2}\Envelope}

\institute{
\textsuperscript{1}Digital Medical Research Center, School of Basic Medical Sciences, Fudan University, Shanghai, China\\
\inst{2}Shanghai Key Laboratory of MICCAI, Shanghai, China\\
\inst{3}Shanghai Institute for Advanced Study of Zhejiang University, Shanghai, China\\
\inst{4}Endoscopy Center and Endoscopy Research Institute, Zhongshan Hospital, Fudan University, Shanghai, China\\
\inst{5}Shanghai Collaborative Innovation Center of Endoscopy, Shanghai, China \\
\inst{6}School of Computer Science and Engineering, Nanjing University of Science and Technology, Nanjing, Jiangsu, China\\
\inst{7}Alliance Manchester Business School, The University of Manchester, Manchester, UK\\
}

\authorrunning{R. Yang et al.}

\maketitle              

\def\thefootnote{\dag}\footnotetext{Equal contribution.}
\def\thefootnote{\Envelope}\footnotetext{Corresponding authors: shuowang@fudan.edu.cn and zhihua.wang@zju.edu.cn}

\begin{abstract}

Determining the necessity of resecting malignant polyps during colonoscopy screen is crucial for patient outcomes, yet challenging due to the time-consuming and costly nature of histopathology examination. While deep learning-based classification models have shown promise in achieving optical biopsy with endoscopic images, they often suffer from a lack of explainability. To overcome this limitation, we introduce EndoFinder, a content-based image retrieval framework to find the 'digital twin' polyp in the reference database given a newly detected polyp. The clinical semantics of the new polyp can be inferred referring to the matched ones. EndoFinder pioneers a polyp-aware image encoder that is pre-trained on a large polyp dataset in a self-supervised way, merging masked image modeling with contrastive learning. This results in a generic embedding space ready for different downstream clinical tasks based on image retrieval.  We validate the framework on polyp re-identification and optical biopsy tasks, with extensive experiments demonstrating that EndoFinder not only achieves explainable diagnostics but also matches the performance of supervised classification models. EndoFinder's reliance on image retrieval has the potential to support diverse downstream decision-making tasks during real-time colonoscopy procedures.


\end{abstract}

\section{Introduction}

Colorectal cancer (CRC) presents a major public health challenge, accounting for approximately 10\% of all cancer incidences worldwide and ranking as the second leading cause of cancer-related deaths \cite{1,2,3}. Colonoscopy stands as the cornerstone for CRC prevention and early detection, primarily through the identification and subsequent management of polyps. During these procedures, clinical endoscopists face critical decisions on whether to remove potentially malignant polyps or opt for active surveillance of benign ones \cite{35}. While the histopathological analysis of biopsied samples serves as the definitive diagnostic method, it is not immediately available during endoscopic examinations. Consequently, clinicians often rely on optical diagnosis through endoscopic imagery for on-the-spot decision-making regarding small colorectal polyps. Artificial intelligence (AI)-based optical diagnosis of polyps has been developed for augmented decision-making during colonoscopy procedures \cite{40}. However, the predominant AI models, characterized by their supervised learning and "black box" nature, suffer from a lack of interpretability. These inductive models demand extensive annotated image datasets for training and need to be re-trained as new data and annotation are acquired, posing significant challenges for scalability and continuous clinical application.

\begin{figure}
    \centering
    \includegraphics[width=1\linewidth]{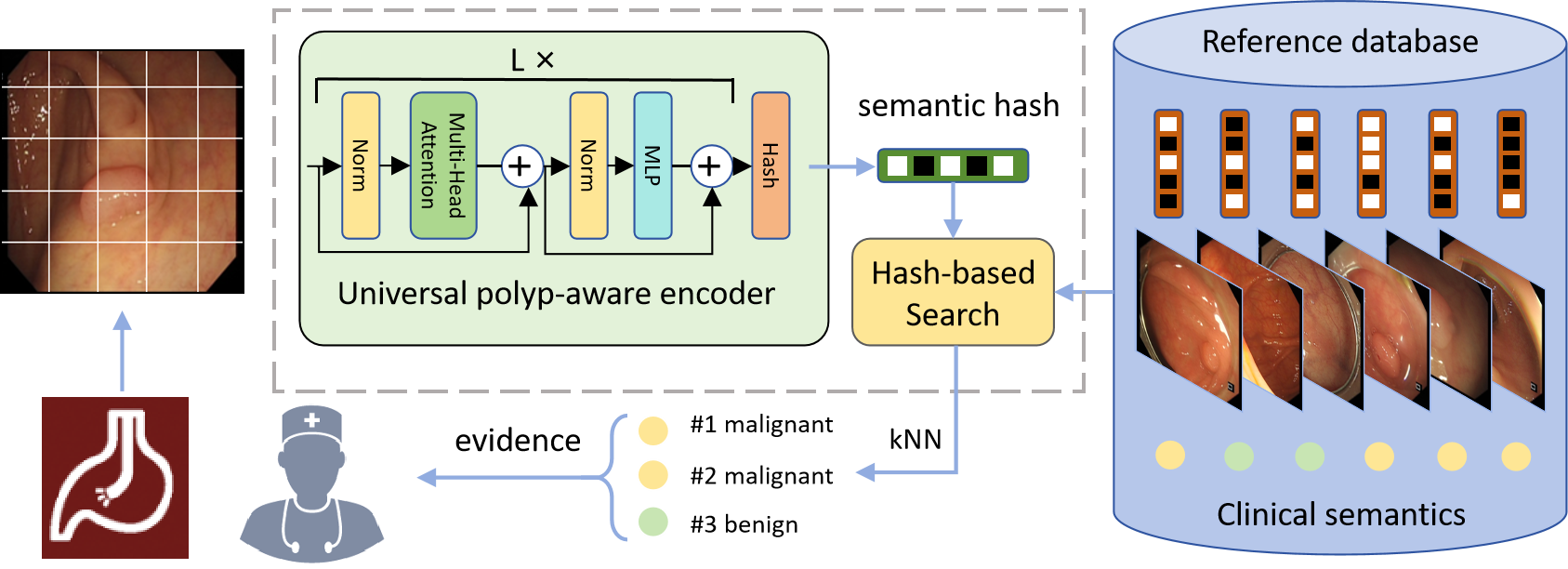}
    \caption{Workflow of the proposed EndoFinder framework. Endoscopic images are encoded into polyp-aware semantic features and discretised into hash codes for fast retrieval. The decision-making is augmented by referring to the historical information of the 'digital twin' polyp in the database. }
    \label{fig1}
\end{figure}

To mitigate the limitations of existing classifiers, we present EndoFinder (Figure \ref{fig1}), an image retrieval framework enhancing diagnostic explainability for colorectal polyps. Inspired by the 'digital twin' concept, EndoFinder identifies a matching 'digital twin' for new polyps in a reference database containing historical data on similar polyps. This approach facilitates interpretable and informed decision-making by leveraging past diagnostic outcomes, offering a scalable solution for real-time polyp diagnosis.

\noindent\textbf{Related work.} Here we review the state-of-the-art performance of optical polyp diagnosis and the medical application of Content-Based Image Retrieval (CBIR). \\
\noindent\textbf{Supervised polyp diagnosis.} Supervised classifiers, particularly those based on deep learning, have matched the expertise of professional endoscopists in optical polyp diagnosis. Ribeiro et al. were pioneers in employing convolutional neural networks (CNNs) for classifying colorectal polyps. Chen et al. developed a system of computer-aided diagnosis system utilizing an Inception v3 architecture to process narrow-band imagery of small colorectal polyps, achieving near-novice doctor accuracy at greater inference speeds. \cite{43}. Yamada et al. developed an AI system based on ResNet152, outperforming expert endoscopists in both internal and external validation \cite{40}. Recently, Krenzer et al. achieved leading accuracy by implementing a method that involves detecting and cropping polyps before classifying them using a Vision Transformer (ViT) \cite{45}. Despite the satisfactory performance of these varied architectural approaches, their clinical applicability is hampered by issues such as limited explainability and vulnerability to data's long-tail distribution.

\noindent\textbf{Content-based image retrieval for medical image analysis.} Unlike inductive methods that derive general rules from the training set, content-based image retrieval presents a transductive alternative to medical image analysis \cite{31,34}. Wang et al. pioneered a CBIR system that facilitates the retrieval of pertinent whole-slide images from vast historical databases \cite{13}. Intrator et al. employed the contrastive learning method SimCLR for polyp representation, advancing polyp video re-identification capabilities \cite{42}. A crucial aspect of CBIR involves constructing an effective embedding space and developing efficient search algorithms for identifying nearest neighbors. For natural images, the focus is increasingly shifting towards learning general and robust representations through self-supervised learning (SSL) on extensive datasets. Pizzi et al. enhanced image copy detection by training CNNs using contrastive learning and score normalization to achieve high-quality embeddings  \cite{17}. Similarly, El-Nouby et al. harnessed Vision Transformer (ViT) networks, integrating InfoNCE with entropy regularizers for improved learning outcomes \cite{11}. To expedite search speeds, Guan et al. devised a method for training CNNs with attention maps to generate semantic hash codes, enabling rapid image retrieval \cite{12}. However, it remains less explored to construct a universal representation for polyp image retrieval.

\noindent\textbf{Contributions.} Our contributions are threefold: Firstly, we propose a novel adaptive self-supervised learning method that merges masked image modeling with contrastive learning to create universal polyp-aware representations, significantly improving the precision of polyp re-identification. Secondly, we introduce an image retrieval approach for explainable polyp diagnosis achieving SOTA performance compared to supervised classifiers. Lastly, we developed a hashing technique to realize real-time image retrieval without accuracy loss.

\section{Methods}

\subsection{Problem Formulation}
Let us denote a task-specific collection of a reference database with clinical semantics as $S = \{(I_i, y_i)\}_{i=1}^N$, where $I_i$ is the image of the $i$-th polyp with clinical categories $y_i \in \{1, 2, \ldots, C\}$ and $N$ is the database size. The task is to infer the clinical label $y_i$ given an image $I_i$ of a newly detected polyp. In general, a supervised classifier uses the reference database to learn the mapping $f_\theta$ parameterized by $\theta$ such that $y_i=f_\theta(I_i)$. Although this method facilitates an end-to-end diagnostic process, it falls short in terms of explainability. 

Drawing inspiration from the K-Nearest Neighbors (KNN) algorithm, the proposed EndoFinder framework builds on the hypothesis that \textit{polyps in close proximity within the embedding space are likely to share similar clinical semantics}. EndoFinder identifies a set of 'digital twins' from the reference database given a test polyp image, leveraging the clinical semantics of the 'digital twins' for transductive reasoning. Formally, the clinical label of a test image can be determined by


\begin{equation}
y_i = \underset{c \in {\{1,...,C\}}}{\mathrm{argmax}}  \sum_{k\in \mathcal{N}(I_i)} \mathbf{1}_{\{y_k = c\}} .
\end{equation}
where $\mathcal{N}(I_i)$  denotes the set of indices corresponding to the K nearest neighbors of the query image $I_i$. Here, $\mathbf{1}_{\{y_k = c\}}$ is the indicator function whether the class label $y_k$ of the $k$th nearest neighbors is equal to the class $c$.

\vspace{-0.3cm}
\subsection{Overview of the EndoFinder design}
\begin{figure}[!h]
    \centering
    \includegraphics[width=1\linewidth]{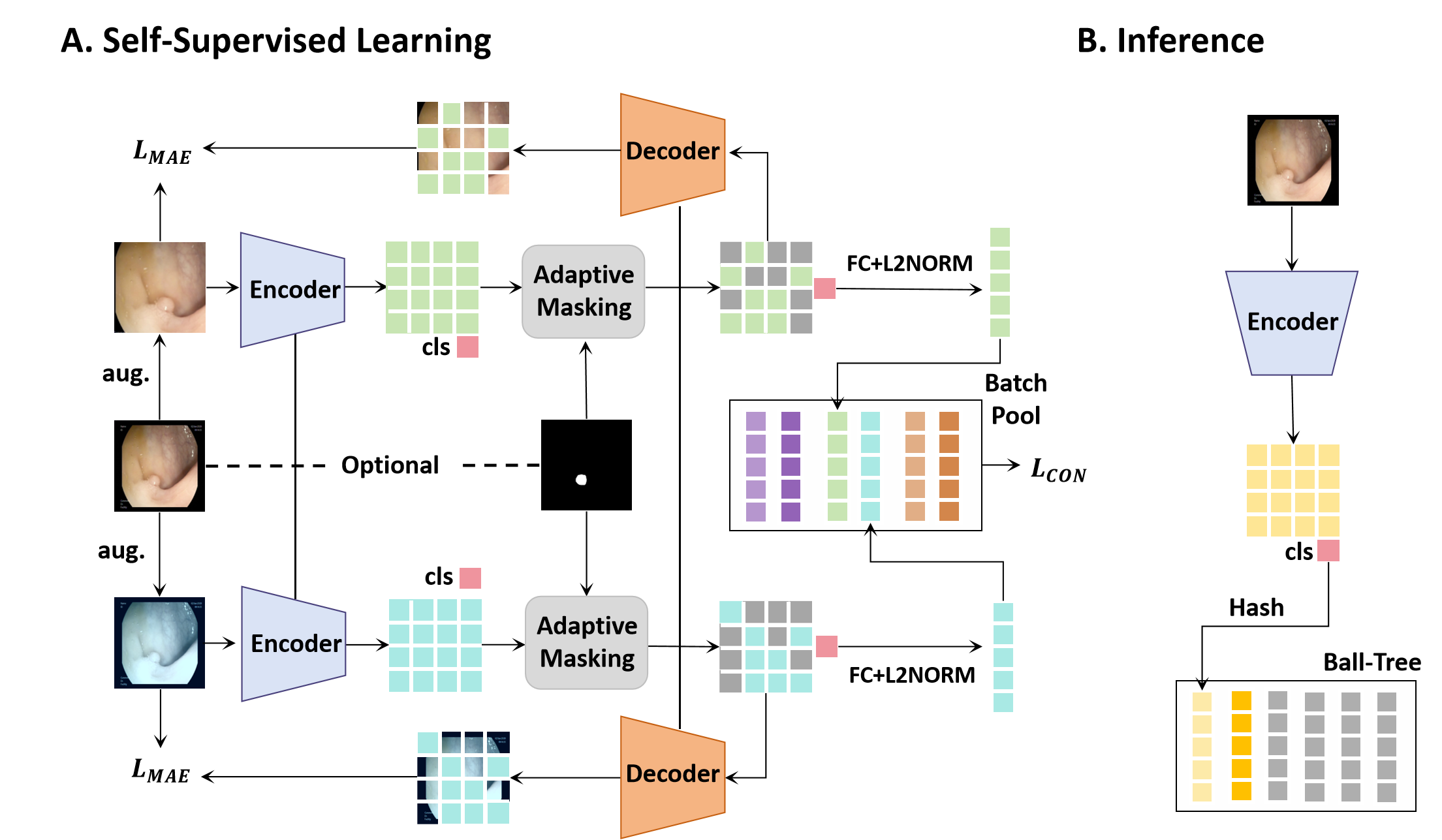}
    \caption{Polyp-aware self-supervised representation learning and inference.}
    \label{fig2}
\end{figure}

The core of EndoFinder is to construct a plausible embedding space for polyp image retrieval, denoted as $z=E_\phi(I)$, where $E_\phi$ represents the feature extractor. Our approach involves learning a universal representation from extensive polyp image datasets through self-supervised learning (SSL) (Figure \ref{fig2}) and subsequently converting this representation into semantic hash codes to enable rapid retrieval.

\noindent\textbf{Universal polyp-aware image encoder:} Drawing inspiration from the effectiveness of masked autoencoder (MAE) and contrastive learning approaches, we integrate these two SSL techniques to pre-train a ViT encoder.

On one hand, the image encoder is trained under the MAE framework to reconstruct masked image patches from the embedding features. In particular, we introduce an adaptive masking strategy that leverages the available polyp segmentation masks. This is realized by masking a larger proportion of background patches compared to foreground patches inversely proportional to the ratio of pixels within the segmentation mask (supplementary material), enabling the encoder to focus on the most informative regions of the image and to generate so-called polyp-aware representation. The MAE reconstruction loss for a batch of N images is the mean square error between the reconstructed image and the original image, focusing solely on the masked regions:
\begin{equation}
L_{MAE} = \frac{1}{2N}\sum_{i=1}^{2N}\frac{1}{|M_i|}\sum_{{k}\in M_i} (\hat{I}_{i,k} - I_{i,k})^2.
\end{equation}
where $M_i$ is a set of non-zero pixels in the masked image $i$,
$\hat{I}_{i,k}$ and $I_{i,k}$ refer to the pixel $k$ in the reconstructed and original image $i$, respectively. 
For a set of $N$ images, we generated $2N$ transformed images through repeated augmentations.

On the other hand, the class token (CLS) from the MAE encoder is subject to a linear projection and L2 normalization, resulting in the embedding feature $z_i \in R^d$. At this stage, contrastive learning is applied, leveraging InfoNCE and Entropy loss to evaluate the distance between augmented images of samples \cite{17}. 
The positive pairs of matching images are $P = \{(i, i+N),(i+N,i)\}_{i \in\{1,...,N\}}$. We denote positive matches for image $i$ as $P_i = \{j|(i,j)\in P\}$. The contrastive InfoNCE loss maximizes the similarity between copies relative to the similarity of non-copies. Entropy loss will push away the nearest neighbor who does not belong to the positive pair. The temperature-adjusted cosine similarity $ s_{i,j} $ is computed between the feature embeddings $ z_i $ and $ z_j $. The loss $L_{CON}$ of contrastive learning is the weighted sum of the infoNCE (first term) and entropy loss (second term), with entropy loss weighted by hyper-parameter $ \gamma$:
\begin{equation}
L_{CON} =- \frac{1}{|P|}\sum_{{(i,j)} \in P}\log\frac{\exp(s_{i,j})}{\sum_{v\neq i} \exp(s_{i,v})} + \gamma \left(-\frac{1}{N} \sum_{i=1}^N \log(\min_{j \notin \hat{P}_i} ||z_i - z_j||) \right) \label{LCON}   
\end{equation}
where $\hat{P_i} = P_i \cup \{i\}$. The overall loss $ L $ is a weighted sum of the aforementioned components, with MAE loss modulated by its weight parameter $ \lambda $:
\begin{equation}
L = L_{CON} + \lambda L_{MAE}.
\end{equation}

\noindent\textbf{Semantic hashing for image retrieval:} To accelerate the image retrieval speed, we transform the features into hash codes through a hashing layer.

The quantization process is defined by:

\begin{equation}
\bar{z}_{i,k} = 
\begin{cases} 
1 & \text{if } {z}_{i,k} \geq 0, \\
-1 & \text{if } {z}_{i,k} < 0.
\end{cases}
\end{equation}
This function assigns a binary code of $1$ if the feature value  $z_{i,k}$ for image $i$ pixel $k$ is non-negative and $-1$ if $z_{i,k}$ is negative. 
Upon obtaining the binary codes, the next step involves retrieving the images most similar to the query image. Using binary codes for constructing a ball tree retrieval system significantly boosts retrieval speed \cite{47}. The retrieval process is based on the similarity of these binary codes to those of the reference images. Once the most similar images are retrieved, a voting mechanism is employed to determine the category of the query image. This mechanism takes into account the categories of the $k$-nearest reference images, thereby leveraging the collective information of the retrieved set for accurate image categorization.

\section{Experiments and Results}
We first train the image encoder on Polyp-18k and then test the utility of EndoFinder in polyp re-identification and optical polyp diagnosis on Polyp-Twin and Polyp-Path, respectively. It is noted that polyps in the datasets do not overlap. We implement two versions of EndoFinder using hashed features (EndoFinder-Hash) or raw features (EndoFinder-Raw). The implementation details and hyper-parameter studies can be found in Supplementary Material.

\vspace{-0.3cm}
\subsection{Datasets}
\noindent\textbf{Polyp-18k:} An in-house dataset of 17,969 polyp images with corresponding polyp segmentation masks for the training of the image encoder of EndoFinder. \\
\noindent\textbf{Polyp-Twin:} A curated set of 200 images representing various angles of 100 distinct polyps (two images for each polyp) from colonoscopy video recordings.\\ 
\noindent\textbf{Polyp-Path:} A dataset of of 147 images with pathological classification \cite{48}. 57\% are malignant and 43\% are benign according.

\subsection{Polyp Re-Identification}
The first task is to retrieve the other paired image of the polyp given one polyp from the Polyp-Twin. We compared our methods to ImageNet pre-trained feature extractors or SSL methods (MAE \cite{16}, ViT-SimCLR and CNN-SimCLR\cite{17}) pre-trained on Polyp-18k.
As evidenced in Table~\ref{tab1}, our model surpasses other models across all metrics.
\begin{table}[h]
\center
\caption{Comparison of Polyp Re-identification Performance. }\label{tab1}
\setlength{\tabcolsep}{6pt}
\begin{tabular}{cc|c|c|c|c|c}
\hline
\multicolumn{1}{l}{}                                                          &             & uAP                             & Acc@1                           & Recall@90\%                     & time(s)                         & FPS                              \\ \hline
\multirow{5}{*}{\begin{tabular}[c]{@{}c@{}}ImageNet \\ Features\end{tabular}} & Resnet50    & 0.365                           & 0.495                           & 0.128                           & 0.485                           & 2.06                             \\
                                                                              & VGG19       & 0.338                           & 0.564                           & 0.118                           & 0.545                           & 1.83                             \\
                                                                              & Densenet121 & 0.377                           & 0.514                           & 0.128                           & 0.471                           & 2.12                             \\
                                                                              & ViT-L16     & 0.243                           & 0.386                           & 0.059                           & 0.451                           & 2.21                             \\
                                                                              & SSCD        & 0.581                           & 0.673                           & 0.326                           & 0.434                           & 2.30                             \\ \hline
\multirow{5}{*}{\begin{tabular}[c]{@{}c@{}}SSL on \\ Polyp-18k\end{tabular}}       & MAE         & 0.470                           & 0.554                           & 0.227                           & 0.453                           & 2.22                             \\ 
                                                                                                                                  & ViT-SimCLR     & 0.591                           & 0.623                           & 0.415                           & 0.454                           & 2.18                             \\                          & CNN-SimCLR        & 0.672                           & \textbf{0.693} & 0.495                           & 0.433                           & 2.30                             \\
                                                                              & EndoFinder-Raw      & \textbf{0.695} & \textbf{0.693} & 0.495                           & 0.456                           & 2.19                             \\
                                                                              & EndoFinder-Hash        & 0.693                           & \textbf{0.693} & \textbf{0.524} & \textbf{0.009} & \textbf{108.57} \\ \hline
\end{tabular}
\end{table}

\begin{figure}[h]
    \centering
    \includegraphics[width=0.85\linewidth]{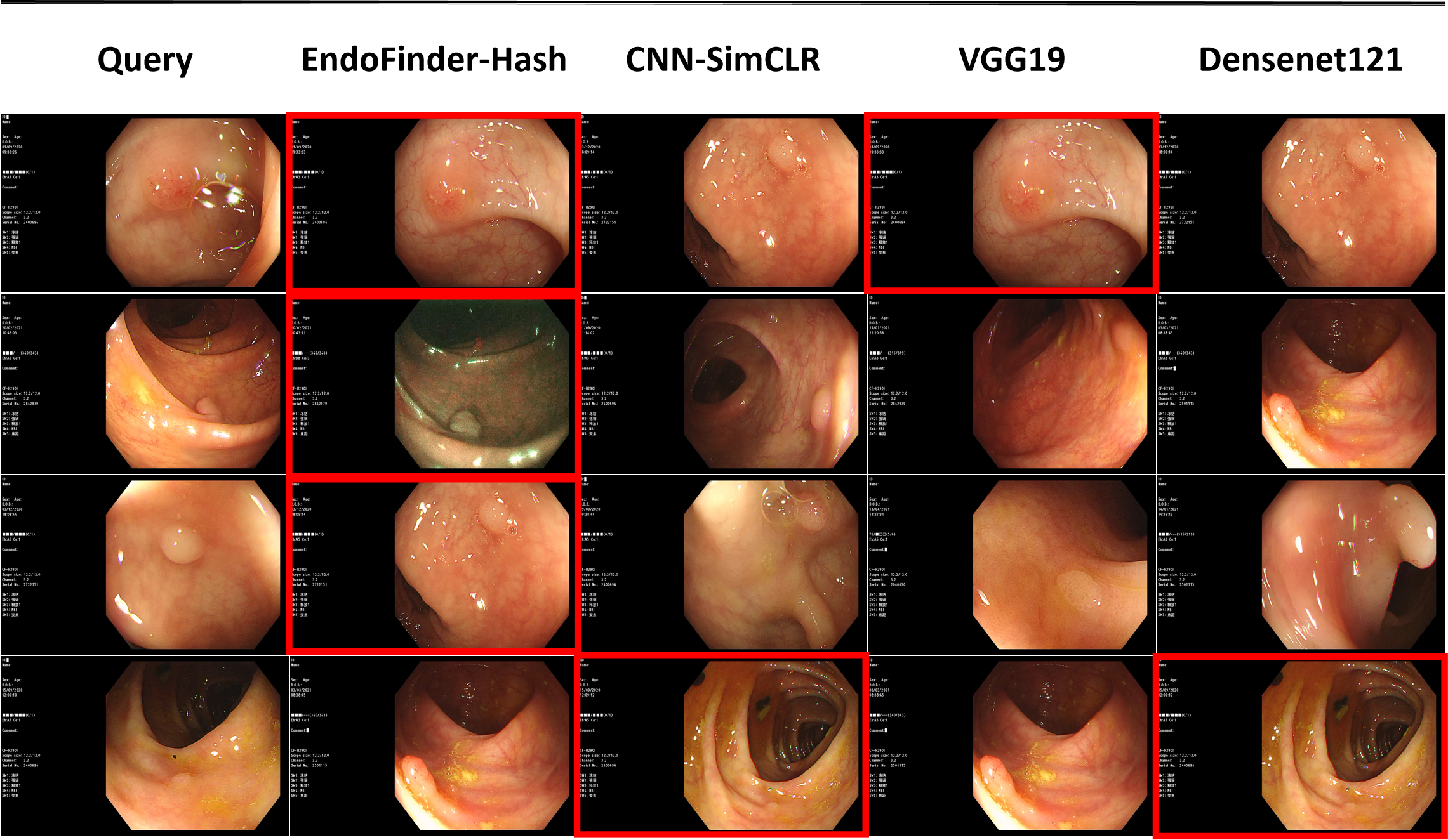}
\caption{Examples of polyp re-identification results. Each row depicts a polyp, showing the query image followed by the first retrieval results from EndoFinder, pre-trained SSCD, VGG19 and Densenet121, respectively. Correct retrievals are bounded in red.}
    \label{fig4}
\end{figure}

Furthermore, We evaluated the speed enhancement achieved using binary codes for image retrieval on a dataset with over 12000 images, as shown in Table~\ref{tab1}. The use of binary codes to construct a ball tree retrieval system significantly enhances retrieval speed. Fig.~\ref{fig4} illustrates a comparative analysis of retrieval outcomes using different feature extractors.

\vspace{-1cm}
\subsection{Optical Polyp Diagnosis} 
\begin{figure}[!h]
    \centering
    \includegraphics[width=0.9\linewidth]{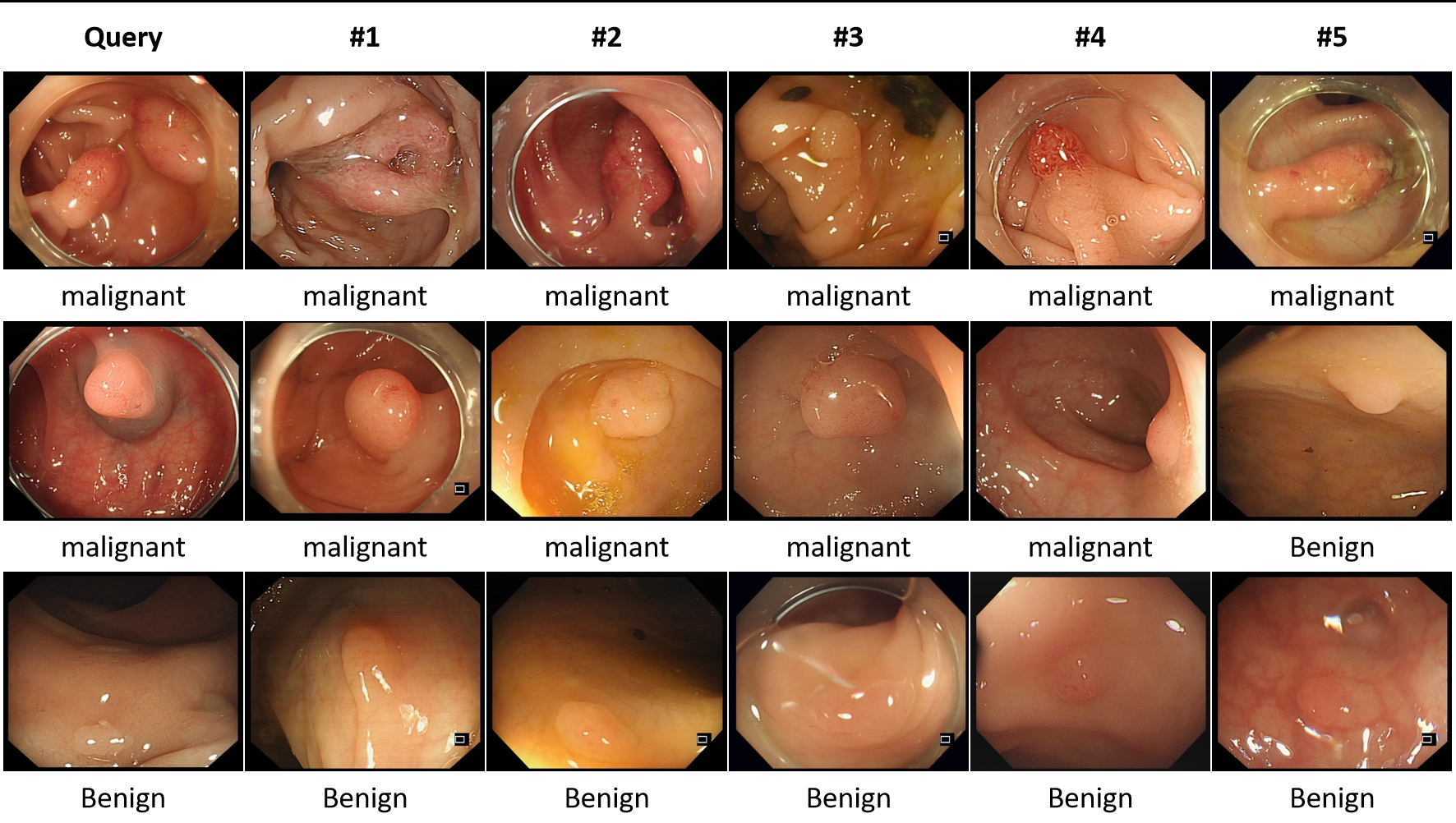}
\caption{Examples of image-retrieval based classification by EndoFinder.}
    \label{fig5}
\end{figure}

After validating the performance of our universal polyp-aware representation, we evaluated the proposed image retrieval-based classification in a more clinically relevant task - determining the pathological malignancy on the Polyp-Path dataset. The outcomes of EndoFinder are illustrated in Fig.\ref{fig5}, demonstrating the model's effectiveness. We compared the performance of image retrieval-based classification using different feature embeddings with supervised classifiers fine-tuned on Polyp-Path with ImageNet pre-trained weights. The performance was evaluated using 5-fold cross-validation, where 4 folds were used as the reference database and the remaining fold was used for testing. The average results are shown in Table~\ref{tab2}.

\vspace{1pt}

\begin{table}[!h]
\center
\caption{Comparison of optical polyp diagnosis performance.
}\label{tab2}
\setlength{\tabcolsep}{8pt}
\begin{tabular}{cc|lllll}
\hline
\multicolumn{1}{c}{}                   &                 & ACC    & SEN    & SPE    & F1     \\ \hline
\multirow{4}{*}{Supervised classifier}        & Resnet50     & 74.482 & 79.095 & 68.988 & 77.880 \\
                                       & VGG19        & 76.550 & 77.954 & 76.259 & 78.848 \\
                                       & Densenet121  & 75.864 & 79.212 & 71.082 & 79.062 \\
                                       & ViT-L16      & 75.862 & 74.286 & \textbf{77.362} & 78.34  \\ \hline
\multirow{4}{*}{\begin{tabular}[c]{@{}c@{}}Retrieval using \\ ImageNet features\end{tabular} }    & Resnet50     & 66.896 & 78.910 & 53.090 & 73.055 \\
                                       & VGG19        & 68.275 & 71.203 & 66.320 & 71.719 \\
                                       & Densenet121  & 73.793 & 80.815 & 64.796 & 77.562 \\
                                       & ViT-L16      & 68.965 & 76.641 & 59.073 & 73.816 \\ \hline
\multirow{4}{*}{\begin{tabular}[c]{@{}c@{}}Retrival using \\ SSL features\end{tabular} } & MAE          & 66.896 & 70.182 & 65.437 & 70.939 \\
                                       & ViT-SimCLR      & 67.586 & 70.753 & 62.588 & 71.593 \\
                                       
                                       & EndoFinder-Raw        & \textbf{77.241} & 81.239 & 73.748 & \textbf{80.445} \\
                                       & EndoFinder-Hash          & 73.793 & \textbf{81.916} & 63.922 & 78.213 \\ \hline
\end{tabular}
\end{table}

\section{Discussion and Conclusion}
By combining advanced SSL techniques, EndoFinder has achieved outstanding performance in polyp image retrieval and pathological classification. Our experimental findings highlight EndoFinder's proficiency in identifying polyp-specific features, as demonstrated by its superior accuracy and F1 scores compared to traditional classification models. Image retrieval performance using EndoFinder features outperforms that of features pre-trained solely through MAE or contrastive learning techniques. This superiority highlights the effectiveness of the adaptive masking strategy and the synergistic benefits of combining SSL techniques.
It should be noted that the EndoFinder features were not fine-tuned on the downstream classification task, demonstrating the power of universal representation learned from large datasets in a self-supervised manner. The polyp-aware semantic hash could serve as a unique identification (UID) to be explored in future studies. By employing hashing-based retrieval methods, EndoFinder ensures scalability to extensive reference datasets. Beyond merely enhancing optical polyp diagnosis performance, EndoFinder has the potential to facilitate various decision-making processes, such as determining the optimal approach for polyp removal by searching and matching similar cases in historical records.

In conclusion, the EndoFinder framework establishes a universal representation for endoscopic images and delivers exceptional performance in real-time polyp diagnosis, complete with explainability.

\noindent\textbf{Disclosure of Interests.} The authors have no competing interests to declare that are relevant to the content of this article.

\noindent\textbf{Acknowledgement.}  This study was supported in part by the
Shanghai Sailing Program (22YF1409300), International Science and Technology Cooperation Program under the 2023 Shanghai Action Plan for Science (23410710400) and National Natural Science Foundation of China (No.62201263).

%
%
%
%

\end{document}